\def\year{2022}\relax
\documentclass[letterpaper]{article} 
\usepackage{aaai22}  
\usepackage{times}  
\usepackage{helvet}  
\usepackage{courier}  
\usepackage[hyphens]{url}  
\usepackage{graphicx} 
\usepackage{natbib}  
\usepackage{caption} 
\DeclareCaptionStyle{ruled}{labelfont=normalfont,labelsep=colon,strut=off} 
\usepackage{algorithm}
\usepackage{algorithmic}

\usepackage{subfigure}
\usepackage{xfrac}
\usepackage{colortbl}
\usepackage{booktabs}  
\usepackage{amsfonts}  
\usepackage{nicefrac}
\usepackage{latexsym}
\usepackage{multirow}
\usepackage{amsmath}
\usepackage{tabularx}
\usepackage{makecell}
\usepackage{xcolor}

\usepackage{newfloat}
\usepackage{listings}
\floatstyle{ruled}
\newfloat{listing}{tb}{lst}{}
\floatname{listing}{Listing}
\title{\textsc{DialogLM}: Pre-trained Model for Long Dialogue \\Understanding and Summarization}

\author{
    Ming Zhong\thanks{Ming completed this work during his internship at Microsoft. Correspondence to: Yang Liu (yaliu10@microsoft.com).}\textsuperscript{\rm 1},
    Yang Liu\textsuperscript{\rm 2},
    Yichong Xu\textsuperscript{\rm 2},
    Chenguang Zhu\textsuperscript{\rm 2},
    Michael Zeng\textsuperscript{\rm 2}
}
\affiliations{
    \textsuperscript{\rm 1}University of Illinois at Urbana-Champaign\\
    \textsuperscript{\rm 2}Microsoft Cognitive Services Research Group\\
    mingz5@illinois.edu, \{yaliu10, yichong.xu, chezhu, nzeng\}@microsoft.com
}

\begin{document}

\maketitle

\begin{abstract}
Dialogue is an essential part of human communication and cooperation. Existing research mainly focuses on short dialogue scenarios in a one-on-one fashion. However, multi-person interactions in the real world, such as meetings or interviews, are frequently over a few thousand words. There is still a lack of corresponding research and powerful tools to understand and process such long dialogues. Therefore, in this work, we present a pre-training framework for long dialogue understanding and summarization. Considering the nature of long conversations, we propose a window-based denoising approach for generative pre-training. For a dialogue, it corrupts a window of text with dialogue-inspired noise, and guides the model to reconstruct this window based on the content of the remaining conversation. Furthermore, to process longer input, we augment the model with sparse attention which is combined with conventional attention in a hybrid manner.  We conduct extensive experiments on five datasets of long dialogues, covering tasks of dialogue summarization, abstractive question answering and topic segmentation. Experimentally, we show that our pre-trained model \textsc{DialogLM} significantly surpasses the state-of-the-art models across datasets and tasks. Source code and all the pre-trained models are available on our GitHub repository\footnote{\url{https://github.com/microsoft/DialogLM}}.

\end{abstract}

\section{Introduction}

Dialogue plays a vital role in interpersonal interaction in daily life, workplace or online forums, and it has drawn extensive attention from both academia and industry~\cite{zhang2020dialogpt}. With the development of speech recognition systems and the growing need of remote work, an increasing number of long conversations are recorded and transcribed, such as meeting minutes, interviews and debates. 
These long dialogues are dense medium of information, bringing challenges for users to quickly understand the gist and extract related information. To address these challenges, many NLP tasks are proposed, including dialogue summarization, dialogue-based question answering and dialogue segmentation~\cite{feng2021survey,feng2021language,zhong2021qmsum,zou2021unsupervised,zou2021topic,chen2021dialsumm,koay2021sliding,hsueh2006automatic}.
However, different from monologic texts like news, long conversations have dialogic structures and lengthy input
which is difficult for current NLP systems to process them.
Thus, exploring a model that can better understand and summarize an entire long dialogue is practically needed.

Recently, pre-trained neural language models achieve remarkable success on a spectrum of natural language tasks~\cite{devlin2018bert,liu2019roberta}. However, these general-purpose models are pre-trained on free-form text data with universal objectives. 
Although this can obtain powerful contextualized language representations, it also limits their ability in specific domains. 
Motivated by this, several dialogue-related pre-trained models have been proposed to tackle different tasks like conversational response generation~\cite{zhang2020dialogpt}, dialogue response ranking~\cite{gao2020dialogue} and multi-party conversation understanding~\cite{gu2021mpc}. Nonetheless, these models are limited to short conversations (e.g, usually fewer than 200 words), and hence are not capable of handling long dialogues (usually longer than 5,000 words) with more speakers and utterances. On the other hand, when it comes to long sequences, subsequent research focuses on improving self-attention method~\cite{DBLP:conf/iclr/KitaevKL20,wang2020linformer} and facilitating the interaction of local and global information~\cite{beltagy2020longformer,DBLP:conf/nips/ZaheerGDAAOPRWY20}. However, these systems are not
designed for dialogues and thus they learn limited knowledge of the dialogic structures.
In general, existing models all have their own dilemmas when dealing with long conversations.

\begin{figure*}
    \centering
    \includegraphics[width=0.57\linewidth]{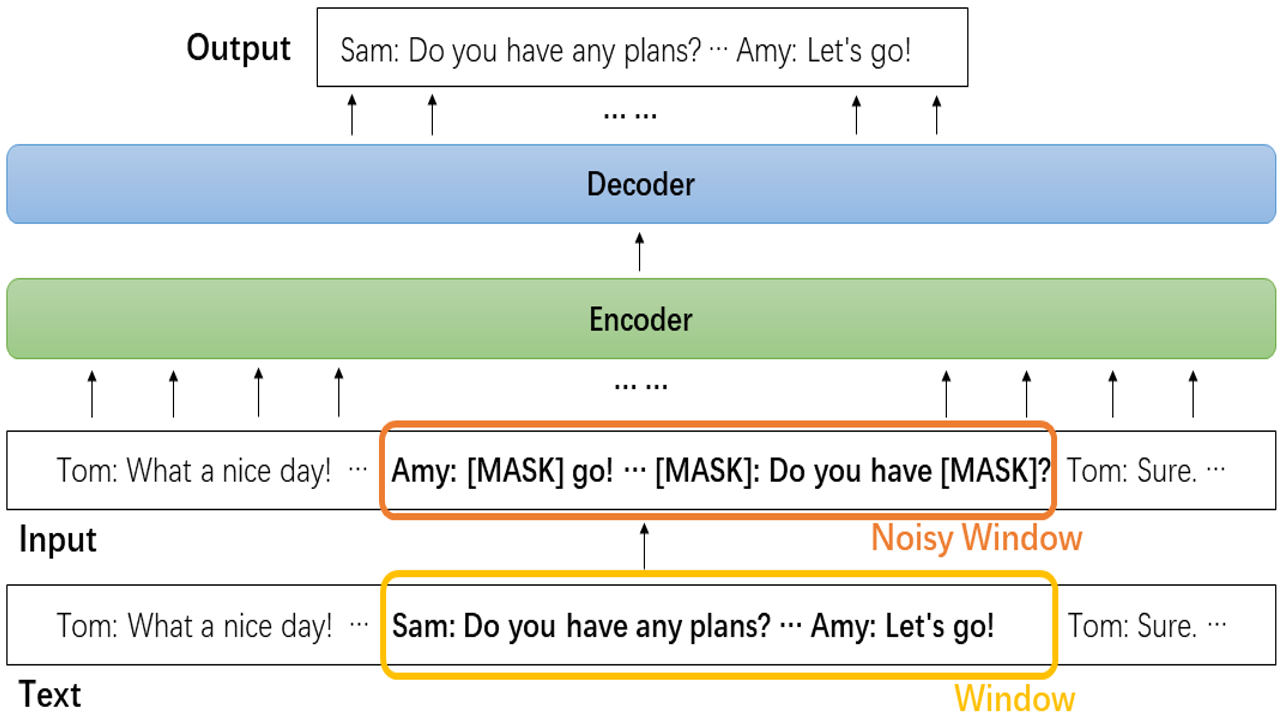}
    \caption{Pre-train task for \textsc{DialogLM}: window-based denoising. We firstly select a window containing multiple turns, and inject different dialogue-inspired noises into it. Finally, we train the model to restore this window based on the noisy window and the rest of the dialogue.}
    \label{fig:window}
\end{figure*}

In this paper, we present \textsc{DialogLM}, a pre-trained neural encoder-decoder model for long dialogue understanding and summarization. \textsc{DialogLM} is established on the sequence-to-sequence model architecture and can be applied to a wide range of natural language processing tasks. As shown in Figure~\ref{fig:window}, we propose a window-based denoising pre-training task on a large dialogue corpus: (1) select a window containing multiple consecutive turns from a conversation; (2) inject arbitrary dialogue-related noise into the window, and (3) train the model to restore this window based on the rest of the conversation. Intuitively, a pre-trained model should be able to reconstruct the noisy window, as speaking style of the interlocutors and topic content exist distributedly in a long conversation.
Compared with sentence-level masking like PEGASUS~\cite{zhang2020pegasus}, a window composed of multiple turns contains more coherent and informative text, which is important for recognizing the format of dialogue. Compared to full-text denoising like BART~\cite{DBLP:conf/acl/LewisLGGMLSZ20}, window-based methods not only require less computational resource, which has significant advantages when dealing with long sequences, but also are better suited to downstream tasks like dialogue summarization.

Furthermore, we design five types of pre-train noises to generate a noisy window based on the characteristics of dialogues: \textit{Speaker Mask}, \textit{Turn Splitting}, \textit{Turn Merging}, \textit{Text Infilling} and \textit{Turn Permutation}. These challenging transformations disrupt both content and order of speakers and utterances.
Therefore, to reconstruct the window,
\textsc{DialogLM} has to fully understand the special format and text style of speaker-utterance pairs, and grasp the general content of the complete dialogue. Moreover, to process longer sequences and reduce training time, we introduce a hybrid attention approach into our model. For most neural layers, we utilize a sparse attention method~\cite{tay2020sparse} to capture  local information; for other layers, global self-attention is used for perceiving the full dialogue semantics.
This hybrid attention approach allows \textsc{DialogLM} to accept more than 8,000 input words while achieving excellent model performance. 

Experimentally, \textsc{DialogLM}  surpasses previous models by a large margin in long dialogue understanding and summarization tasks. Specifically, for dialogue summarization and abstractive question answering, our model outperforms the pre-trained model BART and Longformer~\cite{beltagy2020longformer} on five datasets including meeting and screenplay domains,  achieving new state-of-the-art results across multiple datasets. \textsc{DialogLM} also shows its superiority over strong baseline models for dialogue segmentation task. 
Ablation studies verified the effectiveness of each component in our pre-training framework.
The results demonstrate that each dialogue-inspired noise and the proposed hybrid attention methods can bring further improvements to the model.
In addition to automatic evaluation, for generation tasks, we also perform human evaluation on the generated sequences from three dimensions: fluency, informativeness, and faithfulness to the original dialogue.
Compared with previous powerful models, \text{DialogLM} provides considerable benefits in different perspectives.

\section{Related Work}

\subsection{Pre-trained Neural Models for Dialogues}

Most dialogue-related pre-trained models focus on specific tasks, such as dialogue response generation~\cite{zhang2020dialogpt,bao2020plato,cao2020pretrained}, dialogue response selection~\cite{wu2020tod,gao2020dialogue} and multi-party conversation understanding~\cite{gu2021mpc}. Generally speaking, they either further pre-train general-purpose pre-train models on open-domain conversational data like Reddit for dialogue response generation and selection~\cite{henderson2020convert,zhang2020dialogpt,bao2020plato}, or conduct task-specific training for downstream applications~\cite{ li2020task,wu2020tod,gu2021mpc}. Unlike these previous studies, our pre-train task is not limited to concrete tasks. We hope that through window-based denoising, the model can learn the format and characteristics of dialogues in a general way, thereby performing better in various dialogue-oriented tasks. On the other hand, these work only focus on short dialogue scenes, and usually limit the length of input dialogue. As a result, we still lack powerful NLP tools for long conversations with more speakers and more utterances.

\renewcommand\arraystretch{1.0}
\begin{table*}
\centering \footnotesize
\tabcolsep0.1 in
\small
\begin{tabular}{lll}
\toprule
\textbf{Noise Type} & \textbf{Original Dialogue} & \textbf{Noisy Dialogue} \\
\midrule
\textbf{Speaker Mask} & Tom: The weather is good today! &
\textcolor{blue}{[MASK]}: The weather is good today! \\

\midrule

\textbf{Turn Splitting} & \makecell[l]{Tom: The weather is good today!\\\hspace{2.4em}Do you have any plans?\\\hspace{2.4em}How about we go to play basketball?} &  \makecell[l]{Tom: The weather is good today!\\\textcolor{blue}{[MASK]}:Do you have any plans?\\\textcolor{blue}{[MASK]}:How about we go to play basketball?} \\

\midrule

\textbf{Turn Merging} & \makecell[l]{Tom: The weather is good today!\\\hspace{2.4em}Do you have any plans?\\Bob:\hspace{0.6em}I still have homework to do today.\\\hspace{2.53em}I'm afraid I can't go out to play.} & \makecell[l]{Tom: The weather is good today!\\\hspace{2.4em}Do you have any plans?\\\hspace{2.5em}I still have homework to do today.\\\hspace{2.5em}I'm afraid I can't go out to play.} \\

\midrule

\textbf{Text Infilling} & \makecell[l]{Tom: The weather is good today!\\\hspace{2.4em}Do you have any plans?\\\hspace{2.4em}How about we go to play basketball?} & \makecell[l]{Tom: The weather is \textcolor{red}{[MASK]}\\\hspace{2.4em}Do you have \textcolor{red}{[MASK]} any plans?\\\hspace{2.4em}\textcolor{red}{[MASK]} we go to play basketball?} \\

\midrule

\textbf{Turn Permutation} & \makecell[l]{Tom: Do you have any plans?\\Bob:\hspace{0.4em}How about we go to play basketball?\\Sam:\hspace{0.4em}I still have homework to do today.\\\hspace{2.45em}I'm afraid I can't go out to play.} & \makecell[l]{Sam:\hspace{0.4em}I still have homework to do today.\\\hspace{2.45em}I'm afraid I can't go out to play.\\Tom: Do you have any plans?\\Bob:\hspace{0.4em}How about we go to play basketball?} \\

\bottomrule
\end{tabular}
\caption{Dialogue-related noise for generating a noisy window. The blue [MASK] token means that a speaker name is masked, and the red [MASK] token indicates that a text span in the utterance is masked.}
\label{tab:noise}
\end{table*}

\subsection{Pre-trained  Neural Models for Long Sequences}

Processing long sequences is a  natural need in many NLP tasks.
For the Transformer~\cite{vaswani2017attention} architecture, the core difficulty lies in the computational complexity of the self-attention module, which grows quadratically with the sequence length. 
Recently, many methods are proposed to tackle the long sequence problem by improving the self-attention mechanism.
Specifically, Linformer~\cite{wang2020linformer} uses linear mapping to compress the input sequences under the assumption that the attention mechanism matrix is low-rank. 
Block/bucket-based local attention~\cite{DBLP:conf/iclr/KitaevKL20,DBLP:conf/acl/WangZGCFSCL21,roy2021efficient} utilize the random-projection hashing function or clustering approach to allocate highly similar tokens into a same bucket. 
Sliding window-based attention~\cite{beltagy2020longformer,DBLP:conf/nips/ZaheerGDAAOPRWY20,DBLP:conf/icml/ZhangGSL0DC21} introduce sliding window attention to capture local information and retain part of full attention for global information
We adopt the last two
approaches in this paper to reduce computational cost by mixing Sinkhorn attention~\cite{tay2020sparse} and global attention in the Transformer structure.


\section{Method}

In this section, we first introduce the pre-training task for \textsc{DialogLM}: window-based denoising and five types of dialogue-inspired noises. Then we describe the overall architecture of our pre-trained model.

\subsection{Window-based Denoising}
A long conversation usually involves a core theme and multiple main speakers. For instance, the meetings in the AMI corpus~\cite{carletta2005ami} are about product design in the industrial setting, including discussions among product managers, industrial designers, marketing experts, and user interface designers. Long dialogue with thousands of words can portray the speaking style of different people, e.g., product managers speak actively and energize the audience to help them brainstorm, while marketing experts usually use statistics to state their opinions. Also, a conversation is coherent and its content in different parts are closely related. Therefore, it is possible to infer the speakers and general content of part of the conversation based on the rest context.

Inspired by this, we propose a novel pre-training task for \textsc{DialogLM}: window-based denoising. Formally, given a long dialogue $D = (x_{1}, x_{2}, \cdots, x_{n})$ consisting of $n$ turns, where each turn $x_i$ represents a speaker-utterance pair $x_i = (s_i, u_i)$, we firstly select a random window containing multiple consecutive turns $W = (x_{j}, x_{j+1}, \cdots, x_{j+m})$. Next, we inject several dialogue-related noise into it to generate a noisy window $W' = (x'_{j}, x'_{j+1}, \cdots, x'_{j+m'})$. During the pre-training phase, we concatenate all the turns into a long sequence and replace the window with the noisy version as input to the model, i.e., $X = (x_{1}, \cdots, x'_{j}, \cdots, x'_{j+m'}, \cdots, x_{n})$. The objective is to restore this selected window $W$ by modeling the conditional distribution $p(x_{j}, x_{j+1}, \cdots, x_{j+m}|X)$. As illustrated in Figure~\ref{fig:window}, several turns are chosen as a window, and we generate a noisy window by disrupting their order and masking part of the content and speaker information. The decoder is trained to reconstruct the original window based on the noisy window and the rest of the conversation.

The most relevant work to our proposed pre-train task is full-text denoising by BART~\cite{DBLP:conf/acl/LewisLGGMLSZ20} and sentence-level masking by PEGASUS~\cite{zhang2020pegasus}. However, for sequences with more than 5,000 words, full-text denoising requires unaffordable computational resources. Correspondingly, our window-based approach is a flexible alternative and allows us to add more completely transformed noise without worrying about the model being unable to recover it. On the other hand, unlike documents, a large number of individual turns in a conversation are not informative, such as merely greeting others or chatting about daily routines that have nothing to do with the theme. Therefore, sentence/turn-level masking does not necessarily enable the model to understand the core content of the whole dialogue, but a window with multiple successive turns is more likely to contain meaningful and coherent information. So we argue that, compared with  previous frameworks, window-based denoising can be more suitable for pre-training a model for processing long conversations.

\begin{figure}[t]
    \centering
    \includegraphics[width=1\linewidth]{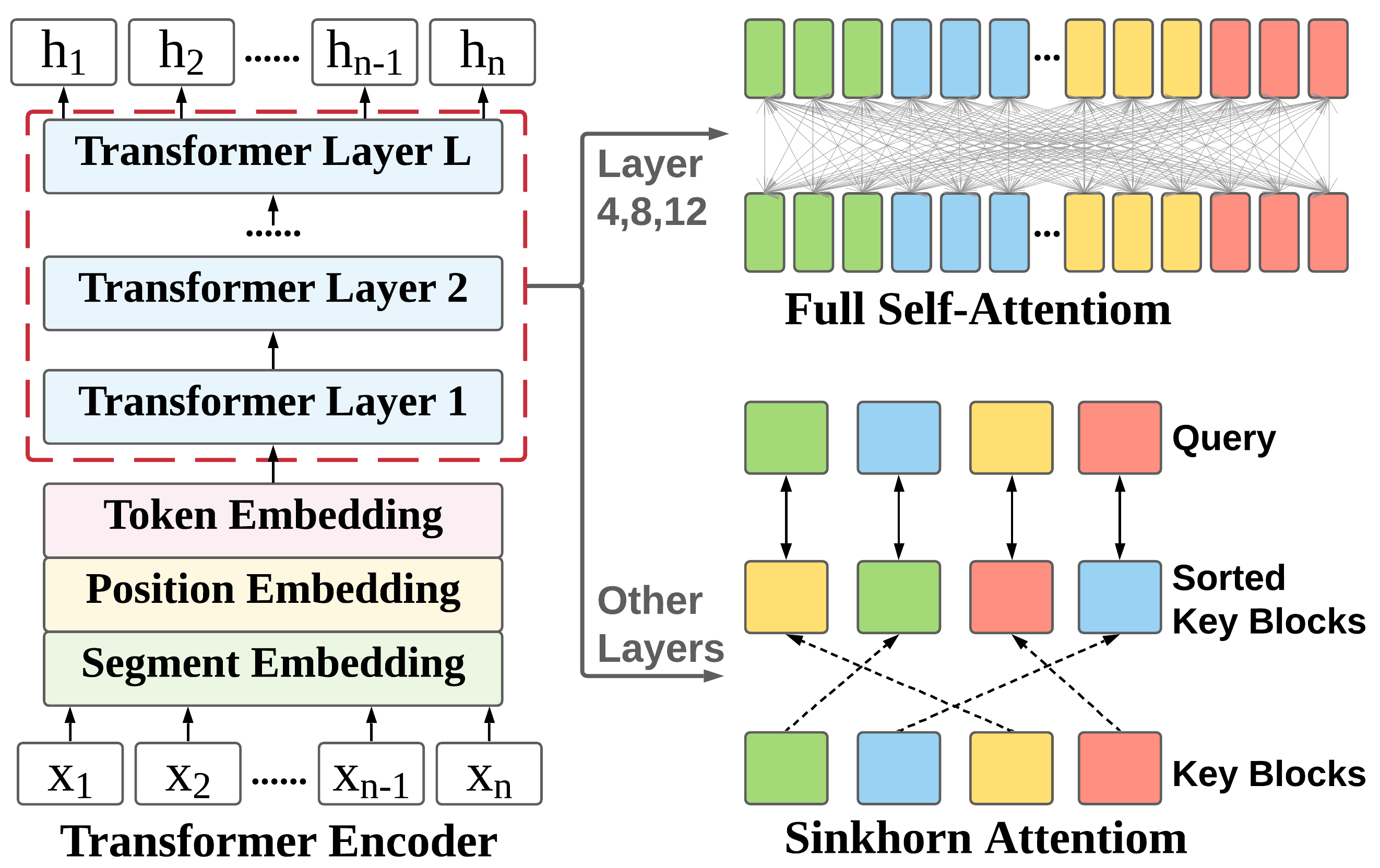}
    \caption{Model architecture for \textsc{DialogLM}. We introduce a hybrid attention approach in Transformer architecture: most layers are equipped with a sparse attention method (Sinkhorn attention) and the rest retain global self-attention.}
    \label{fig:model}
\end{figure}

\subsection{Dialogue-Inspired Noise} The next question is, how do we generate a noisy window? In order to make the model aware of the characteristics of dialogues and their special speaker-utterance format, we design the following five types of noise (see Table~\ref{tab:noise}):

\paragraph{Speaker Mask} For  speaker names of each turn in the window, 50\% of them are randomly sampled and replaced with a special \textsc{[mask\_speaker]} token.

\paragraph{Turn Splitting} A single turn in a conversation can be composed of multiple sentences. We select the turn with the largest number of sentences in the window and split it into multiple turns. We keep the speaker of the first split turn unchanged and take \textsc{[mask\_speaker]} as the speaker of all subsequent newly split turns.

\paragraph{Turn Merging} We randomly sample multiple consecutive turns and merge them into one turn. Keep the speaker of the first turn unchanged, and delete all the speakers of subsequent turns. The number of merged turns is drawn from the Poisson distribution ($\lambda=3$), and is set to be at least 2.

\paragraph{Text Infilling} In a window, we randomly sample several text spans and replace each span with a \textsc{[mask]} token. The length of text span is also drawn from the Poisson distribution ($\lambda=3$). 0-length span corresponds to the insertion of a \textsc{[mask]} token as in \citet{DBLP:conf/acl/LewisLGGMLSZ20}.

\paragraph{Turn Permutation} We shuffle all turns in the window in random order. This noise is  added after \textit{Turn Merging} and \textit{Turn Splitting}. It can further disrupt the speaker and turn information, so that the model can only restore the window when it fully comprehends the context.

\renewcommand\arraystretch{1.0}
\begin{table*}[t]
    \center \footnotesize
    \tabcolsep0.08 in
    \begin{tabular}{lccccccc}
    \toprule
    \textbf{Datasets} & \textbf{\# Dialogues}  & \textbf{\# Turns} & \textbf{\# Speakers} & \textbf{\# Len. of Dialo.} & \textbf{Task} & \textbf{Domain} \\
    \midrule
    \textbf{AMI} & 137 & 535.6 & 4.0 & 5,570.4 & Summarization \& TS & Meeting\\
    \textbf{ICSI} & 59 & 819.0 & 6.3 & 8,567.7 & Summarization \& TS & Meeting \\
    \textbf{QMSum} & 232 & 556.8 & 9.2 & 12,026.3 & Summarization \& QA \& TS & Meeting \\
    \textbf{ForeverDreaming} & 4,348 & 330.7 & - & 7,653.5 & Summarization & Screenplay\\
    \textbf{TVMegaSite} & 22,503 & 327.0 & - & 6,466.6 & Summarization & Screenplay \\
    
    \midrule
    \textbf{MediaSum} & 463,596 & 30.0 & 6.5 & 1,927.2 & Pre-training Data & Interview \\
    \textbf{OpenSubtitles} & 138,086 & 760.0 & - & 5,615.3 & Pre-training Data & TV show \\
    \bottomrule
    \end{tabular}%
    \caption{Statistics of our pre-training data and downstream datasets. QA represents the abstractive question answering task and TS indicates the topic segmentation task.}
  \label{tab:datasets}%
\end{table*}%

\subsection{Model Architecture}

We choose Transformer as our backbone neural architecture because it 
shows promising performance and flexibility on various NLP tasks. For long dialogue processing, pre-trained models based on Transformer
like BART and \textsc{UniLM}~\cite{dong2019unified} have two limitations: 1) they have no pre-training data in dialogue format and no pre-training tasks designed for modeling dialogues; 2) the text length used during pre-training is short (1,024 for BART and 512 for \textsc{UniLM}). Regarding the first issue, we use window-based denoising approach to pre-train our model \textsc{DialogLM} to introduce more dialogue-related knowledge. For the second issue, we leverage hybrid attention method in the Transformer architecture.

Figure~\ref{fig:model} depicts the hybrid attention approach for our model. When dealing with long sequences, encoder self-attention  accounts for the largest computational overhead, so we improve it with the recently proposed sparse Sinkhorn attention~\cite{tay2020sparse, huang2021efficient}. Local attention method such as block-based attention divide the input into several blocks and restricts the words to only attend to the words in their own block. This greatly reduces the computational burden but also loses global information. Sinkhorn attention extends this by additionally introducing  a differentiable sorting network. It sorts the original blocks in a new order, and allows each block to not only  attend to itself, but also to attend to the corresponding block in the new order. As shown in Figure~\ref{fig:model}, the green block can attend to the yellow block because yellow block has the  same position with green block after permutation. With Sinkhorn attention, different layers  learn different permutations, so each block can access information in multiple locations on different  layers. 

However, full dialogue semantics is still indispensable for many applications such as text summarization. Therefore, we keep self-attention of several encoder layers unchanged. In other words, we still use full self-attention in these layers. This hybrid manner enables the interaction of local and global information. Compared with the model that does not introduce sparse attention, it can achieve similar or better performance under the premise of inputting a longer sequence and reducing training time.
It is worth noting that the pre-training task and model modifications we proposed are orthogonal to all Transformer-based pre-trained models. In this paper, we initialize our model with the base version of \textsc{UniLMv2}~\cite{bao2020unilmv2}. And the 4th, 8th and 12th encoder layers of \textsc{UniLMv2} are kept with full self-attention.


\section{Experiments}

\subsection{Implementation Details}

To pre-train \textsc{DialogLM}, we  further train \textsc{UniLM} with the window-based denoising framework for total 200,000 steps on dialogue data, of which 20,000 are warmup steps.
We set batch size to 64 and the maximum learning rate to 2e-5. Pre-training data is the combination of MediaSum dataset~\cite{zhu2021mediasum} and OpenSubtitles Corpus~\cite{lison2016opensubtitles2016}~(see Table~\ref{tab:datasets}). MediaSum is a media interview dataset consisting of 463.6K transcripts. OpenSubtitles\footnote{This corpus is crawled from \url{http://www.opensubtitles.org}} is compiled from a large database of movie and TV subtitles across 60 languages. We use the English part  as the pre-training corpus.
These two large-scale pre-training datasets contain a wealth of long dialogues with  multiple participants and have clear dialogic text structures.
During pre-training, the window size is set to 10\% of the  input length, and the maximum window size is limited to 512 tokens. When generating a noisy window, we first mask 50\% of the speakers, then randomly inject \textit{Turn Splitting} or \textit{Turn Merging}, and utilize \textit{Text Infilling} to mask 15\% tokens in the utterances. Finally, \textit{Turn Permutation} is performed. 8 A100 GPUs with 40GB memory are used to complete the experiments in this paper. All the results listed in this paper are the average of 3 runs. We pre-train two versions of \textsc{DialogLM}: 

\textbf{{\textsc{DialogLM}}} is obtained by further pre-training \textsc{UniLM}-base with the window-based denoising method. Its maximum input length is 5,120 and the tokens exceeding this length is truncated in the experiments.

\textbf{{\textsc{DialogLM}-sparse}} additionally introduces the hybrid attention approach in the pre-training process of \textsc{DialogLM}, so its maximum length is increased to 8,192 tokens.


\renewcommand\arraystretch{0.9}
\begin{table*}[t]
    \centering \footnotesize
    \tabcolsep0.1 in
    \begin{tabular}{lccccccccc}
        \toprule
        \multicolumn{1}{l}{\multirow{3}[1]{*}{\textbf{Models}}}
        & \multicolumn{6}{c}{\textbf{Meeting Summarization}} &
        \multicolumn{3}{c}{\textbf{Abstractive QA}} \\
         & \multicolumn{3}{c}{\textbf{AMI}} & \multicolumn{3}{c}{\textbf{ICSI}} & \multicolumn{3}{c}{\textbf{QMSum}} \\
         & R-1 & R-2 & R-L & R-1 & R-2 & R-L & R-1 & R-2 & R-L\\
         
        \cmidrule(lr){1-1} \cmidrule(lr){2-7} \cmidrule(lr){8-10}
        
        \textsc{PGNet}~($l=2,048$) & 42.60 &  14.01 & \hspace{0.5em}22.62$^*$ & 35.89 & 6.92 & \hspace{0.5em}15.67$^*$ & 28.74 & 5.98 & 25.13 \\
        \textsc{HMNet}~($l=8,192$) & 53.02 & 18.57 & - & 46.28 & 10.60 & - & 32.29 & 8.67 & 28.17 \\
        \textsc{DdaMs}~($l=15,000$) & 53.15 & \textbf{22.32} & \hspace{0.5em}25.67$^*$ & 40.41 & 11.02 & \hspace{0.5em}19.18$^*$ & - & - & - \\
        BART-large~($l=3,072$) & 51.77 & 18.83 & 49.67 & 46.23 & 10.17 & 44.83 & 32.16 & 8.01 & 27.72 \\
        HAT-BART~($l=3,072$) & 52.27 & 20.15 & 50.57 & 43.98 & 10.83 & 41.36 \\
        Longfomer~($l=8,192$) & 54.20 & 20.72 & 51.36 & 43.03 & 12.14 & 40.26 & 31.60 & 7.80 & 20.50 \\
        Longformer-BART-arg~($l=8,192$) & 54.47 & 20.83 & 51.74 & 44.17 & 11.67 & 41.33 & - & - & - \\
        
        \cmidrule(lr){1-1} \cmidrule(lr){2-7} \cmidrule(lr){8-10}
        
        \textsc{UniLM}-base~($l=5,120$) & 51.92 & 18.42 & 49.89 & 46.75 & 11.39 & 45.13 & 29.14 & 6.25 & 25.46 \\
        \textsc{UniLM}-CP~($l=5,120$) & 52.67 & 19.33 & 50.55 & 48.43 & 12.39 & 46.24 & 29.19 & 6.73 & 25.52 \\
        \textsc{DialogLM}~($l=5,120$) & \textbf{54.49} & 20.03 & \textbf{51.92} & 49.25 & 12.31 & 46.80 & \textbf{34.02} & 9.19 & 29.77 \\
        \textsc{DialogLM}-sparse~($l=8,192$) & 53.72 & 19.61 & 51.83 & \textbf{49.56} & \textbf{12.53} & \textbf{47.08} & 33.69 & \textbf{9.32} & \textbf{30.01} \\
        \bottomrule
    \end{tabular}
    \caption{Experimental results on meeting-style datasets. $l$ is the maximum number of input tokens for the corresponding model. Results with $^*$ indicate that ROUGE-L is calculated without sentence splitting.}
    \label{tab:meeting}
\end{table*}

\subsection{Downstream Tasks and Datasets}


After pre-training, we apply \textsc{DialogLM} to three different long dialogue tasks over meeting and screenplay domains, which includes five datasets in total. 



\textbf{Tasks}

\underline{Long Dialogue Summarization}: Given a long conversation ($\textgreater$ 5,000 words), output a concise summary ($\textless$ 512 words) containing its core content.

\underline{Abstractive Question Answering}:  Given a long dialogue and a specific question, generate several sentences as the answer based on  relevant content in the dialogue.

\underline{Topic Segmentation}: Given a long conversation, segment it into multiple parts based on their main topics. Each segment is consist of multiple consecutive utterances. 

\textbf{Datasets}

We employ five popular benchmarks for the above tasks: \texttt{AMI}~\cite{carletta2005ami}, \texttt{ICSI}~\cite{janin2003icsi}, \texttt{QMSum}~\cite{zhong2021qmsum}, \texttt{ForeverDreaming} and \texttt{TVMegaSite}~\cite{chen2021summscreen}. Detailed statistics are given in Table~\ref{tab:datasets}. As shown, these datasets can be divided into two domains: meeting and screenplay.

\underline{Meeting Domain}: \texttt{AMI {\normalfont{and}} ICSI} are meeting transcripts collected from  product design meetings in company and academic group meetings in school, respectively. For each meeting, it contains a meeting summary  and human-annotated topic boundaries. \texttt{QMSum} is a benchmark for query-based multi-domain meeting summarization task. Query type in this dataset can be divided into general query and specific query. The former can be used as the summarization task and the latter can be regarded as abstractive QA task. It also contains human-annotated topic boundaries.

\underline{Screenplay Domain}: \texttt{ForeverDreaming} and \texttt{TVMegaSite} are composed of pairs of TV series transcripts and human-written recaps. They have different dialogue styles from different sources, thus can serve as a challenging testbed for abstractive dialogue summarization.

\subsection{Baselines}

We compare \textsc{DialogLM} with strong baselines as follows:

\textbf{\textsc{UniLM-CP}} refers to the \textsc{UniLM} which is further trained using its original pre-training objective on MediaSum  and OpenSubtitles corpora. The comparison with it can show whether the pre-training framework we proposed is more effective in long conversation scene.

\textbf{Longformer} is a powerful pre-trained model for long sequence processing. We report the results of the Longformer-based model in the previous work on different datasets\footnote{The results of the
Longformer-based model in \texttt{AMI} and \texttt{ICSI} are from \cite{DBLP:conf/acl/FabbriRRWLMR20}, and the results of it in \texttt{QMSum} come from \cite{zhang2021exploratory}. In screenplay domain, the results of Longformer are from \citet{chen2021summscreen}.}. The variant model Longformer-BART-arg~\cite{DBLP:conf/acl/FabbriRRWLMR20} is initialized with the BART-large-CNN\footnote{BART-large-CNN refers to further fine-tuning BART-large on the news summarization dataset CNN/DailyMail.} parameters and uses argument-mining-based source as input.

\textbf{BART} is the state-of-the-art denoising sequence-to-sequence pre-trained model for various generation tasks. We use BART-large in all the experiments. 

\textbf{\textsc{HAT-BART}}~~\cite{rohde2021hierarchical} is a new hierarchical attention Transformer-based architecture that outperforms standard Transformers on several seq2seq tasks.

\textbf{\textsc{HMNet}}~\cite{zhu2020hierarchical} is the state-of-the-art meeting summarization model. It has hierarchical structure and utilizes cross-domain pre-training to recognize the special format of  dialogues.

\textbf{\textsc{DdaMs}}~\cite{feng2020dialogue} is a dialogue discourse-aware summarization model, which utilizes a relational graph encoder to explicitly model the interaction between utterances in a meeting by modeling different discourse relations.

\textbf{Hybrid Model}~\cite{chen2021summscreen} first extracts salient information (up to 1,024 words) from the dialogue and produces summary using BART.

\subsection{Experimental Results}

\paragraph{Results on Meeting Domain}

In this domain, we experiment on two popular summarization datasets \texttt{AMI} and \texttt{ICSI}, and the query-based summarization benchmark \texttt{QMSum}. Since most of the queries are specific on meeting content, \texttt{QMSum} can also be considered as an abstractive question answering task. We use ROUGE~\cite{lin2004rouge} as the evaluation metric. 

As illustrated in Table~\ref{tab:meeting},  \textsc{DialogLM}  achieves state-of-the-art results on most metrics across  datasets. Compared to the backbone model \textsc{UniLM}-base, our proposed pre-train framework brings clear improvements on all the three datasets. Specifically, on \texttt{AMI} and \texttt{ICSI}, \textsc{DialogLM} increases the ROUGE-1 score by more than 2.5 (51.92$\rightarrow$54.49 and 46.75$\rightarrow$49.25), and this improvement reaches to about 5.0 on \texttt{QMSum} (29.14$\rightarrow$34.02. This demonstrates that  window-based denoising approach can assist models to further understand the content of long dialogues, and hence to produce better summaries or answers. However, when returning to the original objective of \textsc{UniLM} (i.e., \textsc{UniLM}-CP), further pre-training does not yield substantial gains. It indicates that  general-purpose pre-training objective limits model's ability to comprehend long dialogues. In comparison with previous state-of-the-art models, the benefits of our model are more pronounced in datasets with longer input text (\texttt{ICSI} and \texttt{QMSum}). For example, \textsc{DialogLM} outperforms \textsc{HMNet} by 2.97 ROUGE-1 points on \texttt{ICSI} and 1.73 R-1 points on \texttt{QMSum}, which indicates that \textsc{DialogLM}, with further pre-training, can be a powerful tool for dealing with long conversation scenarios.

In addition, \textsc{DialogLM}-sparse is able to process longer input with the same memory consumption after introducing hybrid attention mechanism. It  not only reduces the training time, but also performs better on longer meetings. In \texttt{AMI} dataset, since the average length of meetings is about 5,000 (see Table~\ref{tab:datasets}), \textsc{DialogLM} is already competent for  dialogues of this length. As a result, the introduction of hybrid attention has a slightly negative impact on the performance. However, on \texttt{ICSI} and \texttt{QMSum} datasets, as the meeting length exceeds 5,000 words, which is also the common length of one-to-two-hour meetings in real applications, \textsc{DialogLM}-sparse progressively reveals its benefits. It obtains the state-of-the-art ROUGE-2 and ROUGE-L scores on these two datasets, and surpasses \textsc{DialogLM} by about 0.2 points in these metrics. Therefore, the proposed hybrid attention  is genuinely required and can benefit realistic long dialogue scenes.

\renewcommand\arraystretch{0.9}
\begin{table}[t]
    \centering \footnotesize
    \tabcolsep0.08 in
    \begin{tabular}{lcccc}
        \toprule
        \multicolumn{1}{l}{\multirow{2}[1]{*}{\textbf{Models}}}  & \multicolumn{2}{c}{\textbf{AMI}} & \multicolumn{2}{c}{\textbf{QMSum}}\\
         & Pk & Wd & Pk & Wd\\
        
        \cmidrule(lr){1-1} \cmidrule(lr){2-3} \cmidrule(lr){4-5}
        
        \textsc{Random} & 0.47 & 0.65 & 0.52 & 0.70 \\
        \textsc{Even}   & 0.52 & 0.72 & 0.56 & 0.59 \\
        \textsc{UniLM}-base~($l=5,120$) & 0.44 & 0.57 & 0.49 & 0.56 \\
        \textsc{UniLM}-CP~($l=5,120$) & 0.43 & 0.50 & 0.47 & 0.53 \\
        \textsc{DialogLM}~($l=5,120$) & 0.38 & 0.39 & 0.44 & 0.48 \\
        \textsc{DialogLM}-sparse~($l=8,192$) & \textbf{0.36} & \textbf{0.34}  & \textbf{0.38} & \textbf{0.40} \\
        \bottomrule
    \end{tabular}
    \caption{Experimental results on dialogue segmentation task.}
    \label{tab:segmentation}
\end{table}

The results of topic segmentation task on \texttt{AMI} and \texttt{QMSum} are listed in Table~\ref{tab:segmentation}. A long conversation  comprises several segments based on their topics, and  topic segmentation is the task to discover these boundaries. In our experiment, we treat it as a turn-level binary classification task, where a positive label indicates that this turn is the end of a main topic. We follow \citet{liu2019text} to insert a special token [\textsc{cls}] to the start of each each turn, and utilize the hidden state of [\textsc{cls}]  as turn-level representation for classification.

We use standard metrics \textit{Pk}~\cite{beeferman1999statistical} and \textit{WinDiff}~\cite{pevzner2002critique} to evaluate segmentation models. Lower scores of these two metrics indicate that the predicted segmentation is closer to the ground truth. \textsc{Random} is the baseline that randomly selects boundary points throughout the conversation, while \textsc{Even} segments the whole dialogue evenly. The results on two datasets show the same trend: dialogue-related pre-training can boost performance, and the hybrid attention augmentation could further improve upon this.

\paragraph{Results on Screenplay Domain} In this domain, plot details emerge indirectly in character dialogues and are scattered throughout the full transcript. To succinct plot summary,  the model needs to locate and incorporate these aspects in the long dialogue. 

As shown in Table~\ref{tab:summscreen}, similar to the meeting domain, our two models achieve superior performance across the majority of criteria. On both \texttt{ForeverDreaming} and \texttt{TVMegaSite}, \textsc{DialogLM}-sparse outperforms BART-Large by 1.93 and 2.04 ROUGE-1 points (33.82$\rightarrow$35.75 and 43.54$\rightarrow$45.58), respectively. This improvement becomes more prominent when compared to \textsc{UniLM}-base which has no dialogue-oriented pre-training and sparse attention. With merely dialogue pre-training data and no dialogue-specific pre-training framework, \textsc{UniLM}-CP can be enhanced but still  inferior to  \textsc{DialogLM}. On the other hand, \textsc{DialogLM}-sparse consistently outperforms \textsc{DialogLM} in these datasets. We believe this is due to the relatively long length of screenplay (over 6,000 words, see Table~\ref{tab:datasets}), and the tail part of  its transcript is still critical. Models that can deal with longer sequences, such as \textsc{DialogLM}-sparse, are able to capture more dialogue content and storylines,  leading to further improvement in a variety of circumstances.

\renewcommand\arraystretch{0.9}
\begin{table}[t]
    \centering \footnotesize
    \tabcolsep0.05 in
    \begin{tabular}{lccccccccc}
        \toprule
        \multicolumn{1}{l}{\multirow{2}[1]{*}{\textbf{Models}}}  & \multicolumn{3}{c}{\textbf{ForeverDreaming}} & \multicolumn{3}{c}{\textbf{TVMegaSite}} \\
         & R-1 & R-2 & R-L & R-1 & R-2 & R-L \\
         
        \cmidrule(lr){1-1} \cmidrule(lr){2-4} \cmidrule(lr){5-7}
        
        Longformer & 25.90 & 4.20 & 23.80 & 42.90 & \textbf{11.90} & 41.60 \\
        Hybrid Model & 25.30 & 3.90 & 23.10 & 38.80 & 10.20 & 36.90 \\
        BART-large & 33.82 & 7.48 & 29.07 & 43.54 & 10.31 & 41.35 \\
        
        \cmidrule(lr){1-1} \cmidrule(lr){2-4} \cmidrule(lr){5-7}
        
        \textsc{UniLM}-base & 32.16 & 5.93 & 27.27 & 43.42 & 9.62 & 41.19 \\
        \textsc{UniLM}-CP & 33.29 & 6.74 & 28.21 & 44.07 & 9.96 & 41.73 \\
        \textsc{DialogLM} & 35.42 & 8.23 & 30.61 & 45.04 & 10.45 & 42.71 \\
        \textsc{DialogLM}-sparse & \textbf{35.75} & \textbf{8.27} & \textbf{30.76} & \textbf{45.58} & 10.75 & \textbf{43.31} \\
        \bottomrule
    \end{tabular}
    \caption{Experimental results on screenplay-style datasets: ForeverDreaming and TVMegaSite.}
    \label{tab:summscreen}
\end{table}

\paragraph{Ablation Study} To better understand the contribution of each component in our pre-trained model, we conduct comprehensive ablation studies on \texttt{QMSum} and \texttt{TVMegaSite}, which can be viewed as representatives of the meeting and screenplay domain respectively. Overall, our proposed pre-training framework, i.e. window-based denoising, greatly strengthens the capacity of the general-purpose pre-trained model to process long conversations. It is reflected in Table~\ref{tab:Ablation}: removing ``Pre-train'' results in substantial performance degradation on both datasets. Furthermore, all five dialog-inspired noises contribute to the pre-training process. The most important of these are \textit{Turn Split} and \textit{Turn Merging}. We think this is because the model can not denoise them without being aware of the dialogic structure and the main content. \textit{Speaker Mask} brings the least benefit, due to the restoration of other noises also implicitly requires the model to figure out who is the interlocutor of each turn. Additionally, whether to introduce the hybrid attention mechanism allows our model to be flexibly applicable to more scenarios, so we release two versions of \textsc{DialogLM} to accommodate the circumstance of varying durations of dialogues.

\paragraph{Human Evaluation} As factual inconsistency between the source document and the generated sequence is critical in the dialogue domain~\cite{zhong2021qmsum}, it is indispensable to evaluate the abstractive model manually. Specifically, we ask 10 graduate students to rank four different summaries (generated by \textsc{UniLM}-base, BART, \textsc{DialogLM}-sparse and reference summary) according to three metrics: fluency, informativeness and faithfulness to the original dialogue. Ranking first means the best performance on this metric. We randomly select 30 samples from each test set of \texttt{QMSum} and \texttt{TVMegaSite} for human evaluation. The results are provided in Table~\ref{tab:human_evaluation}. From the perspective of fluency, after further pre-training, \textsc{DialogLM} can output more coherent sentences than \textsc{UniLM}, and it is comparable to BART. However, the performance of all neural models is still far from the human-annotated answers or summaries. In terms of the other two metrics, \textsc{DialogLM} is substantially more informative and reliable when comparing to the prior state-of-the-art model BART. The capacity of \textsc{DialogLM} to better grasp the structure and content of conversations is largely responsible for this improvement. Regarding the reference summary, it obtains high scores on \texttt{QMSum}, but performs poorly on the faithfulness of \texttt{TVMegaSite}. This is because for screenplay domain, some useful information is displayed in the video rather than in the dialogue transcript. For example, if two people are eating in a restaurant, we can easily see this from the video of the TV series, but this may not be present explicitly in the transcript. It leads to the fact that the reference summaries written by human can be  informative but contain details that are not visible in the conversation, which are regarded as unfaithful to the original dialogue.

\renewcommand\arraystretch{0.9}
\begin{table}[t]
    \centering \footnotesize
    \tabcolsep0.1 in
    \begin{tabular}{lcc}
        \toprule
        \textbf{Model}  & \textbf{QMSum} & \textbf{TVMegaSite} \\
        \midrule
        \textsc{DialogLM}-sparse & 33.69 & \textbf{45.58} \\
        \quad - Sparse Attention & \textbf{34.02} & 45.04 \\
        \quad - Pre-train & 29.14 & 43.42 \\
        \quad - Speaker Mask & 33.52 & 45.31 \\
        \quad - Turn Splitting~/~Merging & 32.76 & 44.23  \\
        \quad - Text Infilling & 33.27 & 44.79 \\
        \quad - Turn Permutation & 33.22 & 44.64 \\
        \bottomrule
    \end{tabular}
    \caption{Ablation study of \textsc{DialogLM} (ROUGE-1 score). `-’ means we remove the module from the original model.} 
    \label{tab:Ablation}
\end{table}

\renewcommand\arraystretch{0.9}
\begin{table}[t]
    \centering \footnotesize
    \tabcolsep0.07 in
    \begin{tabular}{lccccccccc}
        \toprule
        \multicolumn{1}{l}{\multirow{2}[1]{*}{\textbf{Models}}}  & \multicolumn{3}{c}{\textbf{QMSum}} & \multicolumn{3}{c}{\textbf{TVMegaSite}} \\
         & Ful. & Info. & Faith. & Ful. & Info. & Faith. \\
         
        \cmidrule(lr){1-1} \cmidrule(lr){2-4} \cmidrule(lr){5-7}

        \textsc{UniLM} & 3.20 & 3.07 & 3.13 & 2.83 & 3.23 & 2.63 \\
        BART & 2.67 & 3.27 & 3.43 & \textbf{2.43} & 2.73 & 2.50 \\
        \textsc{DialogLM} & \textbf{2.43} & \textbf{2.37} & \textbf{2.20} & 2.53 & \textbf{2.37} & \textbf{2.13} \\
        
        \cmidrule(lr){1-1} \cmidrule(lr){2-4} \cmidrule(lr){5-7}
        
        Ref. Sum. & 1.70 & 1.30 & 1.23 & 2.20 & 1.67 & 2.73 \\
        \bottomrule
    \end{tabular}
    \caption{Results of human evaluation by ranking. Ful., Info., and Faith. represent fluency, infomativeness and faithfulness to the original dialogue, respectively. Ref. Sum. refers to the human-annotated reference summary.}
    \label{tab:human_evaluation}
\end{table}

\renewcommand\arraystretch{0.9}
\begin{table*}[t]
    \centering \footnotesize
    \tabcolsep0.1 in
    \begin{tabular}{lccccccccc}
        \toprule
        \multicolumn{1}{l}{\multirow{3}[1]{*}{\textbf{Models}}}
        & \multicolumn{6}{c}{\textbf{Meeting Summarization}} &
        \multicolumn{3}{c}{\textbf{Abstractive QA}} \\
         & \multicolumn{3}{c}{\textbf{AMI}} & \multicolumn{3}{c}{\textbf{ICSI}} & \multicolumn{3}{c}{\textbf{QMSum}} \\
         & R-1 & R-2 & R-L & R-1 & R-2 & R-L & R-1 & R-2 & R-L\\
         
        \cmidrule(lr){1-1} \cmidrule(lr){2-7} \cmidrule(lr){8-10}
        
        \textsc{DialogLM}-best & 54.49 & 20.03 & 51.92 & 49.56 & 12.53 & 47.08 & 34.02 & 9.32 & 30.01 \\
        
        \cmidrule(lr){1-1} \cmidrule(lr){2-7} \cmidrule(lr){8-10}
        
        \textsc{LED}-large~($l=5,120$) & 53.52 & 19.28 & 51.08 & 48.98 & 12.38 & 46.30 & 33.70 & 9.61 & 29.88 \\
        \textsc{DialogLED}-large~($l=5,120$) & \textbf{54.80} & \textbf{20.37} & \textbf{52.26} & \textbf{50.11} & \textbf{13.23} & \textbf{47.25} & \textbf{34.50} & \textbf{9.92} & \textbf{30.27} \\
        \bottomrule
    \end{tabular}
    \caption{Experimental results of DialogLED on meeting-style datasets. \textsc{DialogLM}-best indicates the best results of different versions of DialogLM on these datasets.}
    \label{tab:meeting_led}
\end{table*}

\renewcommand\arraystretch{0.9}
\begin{table}[t]
    \centering \footnotesize
    \tabcolsep0.05 in
    \begin{tabular}{lccccccccc}
        \toprule
        \multicolumn{1}{l}{\multirow{2}[1]{*}{\textbf{Models}}}  & \multicolumn{3}{c}{\textbf{ForeverDreaming}} & \multicolumn{3}{c}{\textbf{TVMegaSite}} \\
         & R-1 & R-2 & R-L & R-1 & R-2 & R-L \\
         
        \cmidrule(lr){1-1} \cmidrule(lr){2-4} \cmidrule(lr){5-7}
        
        \textsc{DialogLM}-best & 35.75 & 8.27 & 30.76 & \textbf{45.58} & 10.75 & \textbf{43.31} \\
        
        \cmidrule(lr){1-1} \cmidrule(lr){2-4} \cmidrule(lr){5-7}
        
        \textsc{LED}-large & 35.47 & 8.13 & 30.28 & 44.53 & 10.91 & 42.06 \\
        \textsc{DialogLED}-large & \textbf{36.70} & \textbf{8.68} & \textbf{31.38} & 45.22 & \textbf{11.69} & 42.86 \\
        \bottomrule
    \end{tabular}
    \caption{Results of DialogLED on screenplay-style datasets: ForeverDreaming and TVMegaSite. \textsc{DialogLM}-best indicates the best results of different versions of DialogLM on these datasets.}
    \label{tab:summscreen_led}
\end{table}

\section{DialogLM with LED}

To verify whether our proposed framework can be used for various types of pre-trained models, we also experiment on the Longformer-Encoder-Decoder (LED)~\cite{beltagy2020longformer}. Specifically, we choose LED as the model architecture and initialize it with the original weights, and further train it on the same long dialogue data using window-based denoising task, resulting in a model we call DialogLED. As shown in Table~\ref{tab:meeting_led} and \ref{tab:summscreen_led}, although LED is already a powerful baseline, the introduction of the pre-training framework in this paper brings remarkable improvements on all five datasets. Meanwhile, since DialogLED is a large-version model with more parameters, it beats DialogLM on all four datasets except \texttt{TVMegaSite}, achieving the new state-of-the-art results. These experiments demonstrate that the approach proposed in this paper can be widely used with pre-trained models of different model structures and sizes, and persistently assist the models to better understand the content of long conversations. To meet the needs of processing long dialogues in different scenarios, we release all the different versions of DialogLM and DialogLED in our GitHub repository\footnote{\url{https://github.com/microsoft/DialogLM}}.

\section{Conclusion}

In this paper, we propose a novel pre-training framework for long dialogue understanding and summarization. In particular, given a long conversation, we substitute a portion of it with a noisy window comprising five dialogue-inspired noises, and let the model generate the original dialogue window. As a consequence, the pre-trained model can efficiently realize the dialogic structure and capture the essential information, allowing it to reconstruct any section of the conversation. Moreover, we present a hybrid attention approach to adapt to longer dialogue scenarios. Experiments show that our pre-trained model \textsc{DialogLM} outperforms the previous state-of-the-art models on five benchmarks with three long dialogue understanding and summarization tasks.

\section{Acknowledgements}
We thank Shaohan Huang for providing  guidance on pre-training UniLM-CP. We would also like to thank annotators such as Liang Chen for their hard work and reviewers for their valuable comments. 

\bibliography{aaai22}

\end{document}